%% file: paper.tex
\documentclass{article}
\usepackage{xcolor}
\usepackage[preprint]{neurips_2026}

\usepackage[utf8]{inputenc} 
\usepackage[T1]{fontenc}    
\usepackage{hyperref}       
\usepackage{url}            
\usepackage{booktabs}       
\usepackage{amsfonts}       
\usepackage{nicefrac}       
\usepackage{microtype}      
\usepackage{xcolor}         
\usepackage{multirow}
\usepackage{enumitem}
\usepackage{graphicx}
\usepackage{amsmath}
\usepackage{array}
\usepackage{tabularx}
\newcolumntype{Z}{>{\centering\arraybackslash}X}
\newcommand{\twometrics}[2]{%
  \makebox[\linewidth][c]{%
    \makebox[0.6\linewidth][c]{#1}%
    \makebox[0.6\linewidth][c]{#2}%
  }%
}
\newcommand{\ourmethod}{\textsc{Sugar}~}

\title{\textsc{Sugar}: A \underline{S}calable H\underline{u}man-Video-Driven \underline{G}eneralizable Hum\underline{a}noid Loco-Manipulation Lea\underline{r}ning Framework}

\author{
Tianshu Wu\textsuperscript{1*} \quad
Xiangqi Kong\textsuperscript{2*} \quad
Yue Chen\textsuperscript{1*} \quad \\
Qize Yu\textsuperscript{1} \quad
Hang Ye\textsuperscript{1} \quad
Jia Li\textsuperscript{1} \quad
Yizhou Wang\textsuperscript{1} \quad
Hao Dong\textsuperscript{1$\dagger$} \quad \\
$^1$CFCS, School of Computer Science, Peking University \quad \\
$^2$School of Computer Science and Engineering, Beihang University \\[0.3em]
}

\begin{document}

\maketitle
\input{tex/0_abs}
\input{tex/1_intro}
\input{tex/2_related_work}
\input{tex/3_method}
\input{tex/4_experiment}
\input{tex/5_conclusion}

\bibliographystyle{plainnat}
\bibliography{references}

\appendix
\input{tex/6_appendices}




\end{document}

%% file: tex/0_abs.tex
\begin{abstract}
Building humanoid robots that perform generalizable whole-body loco-manipulation in the real world remains a fundamental challenge: existing approaches either rely on heavy task-specific reward engineering, rigidly replay reference motions that fail to generalize, or depend on costly teleoperation that limits scalability.
While human videos capture diverse human behaviors, the motion priors inferred from them are inherently imperfect, suffering from occlusion, contact artifacts, and retargeting errors that render them unsuitable for direct policy learning.
To this end, we present \textsc{Sugar}, a data-driven framework that converts diverse human videos into deployable humanoid loco-manipulation skills, without any task-specific reward engineering or reference-motion conditioning at inference.
\ourmethod proceeds in three stages:
First, a fully automated pipeline extracts kinematic interaction priors including human-object motion trajectories and contact labels from diverse human videos.
Second, a privileged physics-based refiner utilizes a unified mimic-style reward and a progressive state pool to transform imperfect kinematic interaction priors into physically feasible, high-fidelity skills.
Third, the refined skills are distilled into a deployable autonomous policy, which is implemented as a command generator paired with a command tracker.
We evaluate our method on six representative loco-manipulation tasks in both simulation and real-world humanoid hardware. \ourmethod substantially outperforms reference-tracking baselines, and its performance scales clearly with the amount of human video data.
It also achieves zero-shot real-world transfer with reliable closed-loop execution, autonomous failure recovery and stable long-horizon performance under external perturbations. Project Page: \href{https://tianshuwu.github.io/sugar-humanoid/}{{https://tianshuwu.github.io/sugar-humanoid/}}
\end{abstract}

%% file: tex/1_intro.tex
\section{Introduction}

A general-purpose humanoid assistant must seamlessly coordinate locomotion, balance, and contact-rich object manipulation in unstructured environments. Existing approaches each face a scalability bottleneck. Reinforcement learning from scratch achieves remarkable single-task results~\citep{xue2025openingsimtorealdoorhumanoid, he2025viralvisualsimtorealscale, liu2024visualwholebodycontrollegged, wang2025learningvisiondrivenreactivesoccer, su2025realworldcooperativecompetitivesoccer} but relies on heavy task-specific reward engineering and environment design. Reference-motion tracking~\citep{zhao2025resmimicgeneralmotiontracking, weng2025hdmilearninginteractivehumanoid} attains high-fidelity behavior but rigidly binds the policy to recorded trajectories, limiting generalization across object geometries and configurations. Teleoperation-based imitation learning~\citep{luo2025sonicsupersizingmotiontracking, ze2025twist2scalableportableholistic, li2025cloneclosedloopwholebodyhumanoid, ben2025homiehumanoidlocomanipulationisomorphic, li2025amoadaptivemotionoptimization} produces high-quality embodiment-consistent data but demands extensive human effort and specialized hardware. Across all three paradigms, the data and engineering costs grow steeply with task diversity, hindering progress toward general-purpose interaction.

Diverse human videos~\citep{wang2026humanxagilegeneralizablehumanoid,mao2024learningmassivehumanvideos, yang2026zerowbclearningnaturalvisuomotor,weng2025hdmilearninginteractivehumanoid} offer a compelling escape from this bottleneck. However, while human-object interaction (HOI) videos are abundant, the kinematic data extracted from them is inherently imperfect. Severe occlusion, contact artifacts, and retargeting errors render this data physically implausible for direct imitation. Consequently, current methods either strictly focus on object-free locomotion~\citep{zhao2025resmimicgeneralmotiontracking, he2024learninghumantohumanoidrealtimewholebody, ji2025exbody2advancedexpressivehumanoid}, or rigidly replay recorded HOI trajectories without generalizing to novel configurations~\citep{weng2025hdmilearninginteractivehumanoid, zhao2025resmimicgeneralmotiontracking} or surviving the sim-to-real gap on physical hardware~\citep{xu2026intermimicuniversalwholebodycontrol, tessler2024maskedmimicunifiedphysicsbasedcharacter}. To date, no framework provides a pathway from scalable video to reference-free loco-manipulation on real hardware.

We present \textsc{Sugar}, a data-driven framework that bridges this gap. Our key insight is that imperfect video-extracted data, despite its noise and artifacts, captures coarse but complete task logic, the rough body trajectories, contact events, and object motions that together define what an interaction is trying to accomplish. While too noisy for direct imitation, this data can be progressively refined into physically grounded training signals through simulation, and distilled into an autonomous policy.

\begin{figure}[t]
    \centering
    \includegraphics[width=1.0\textwidth]{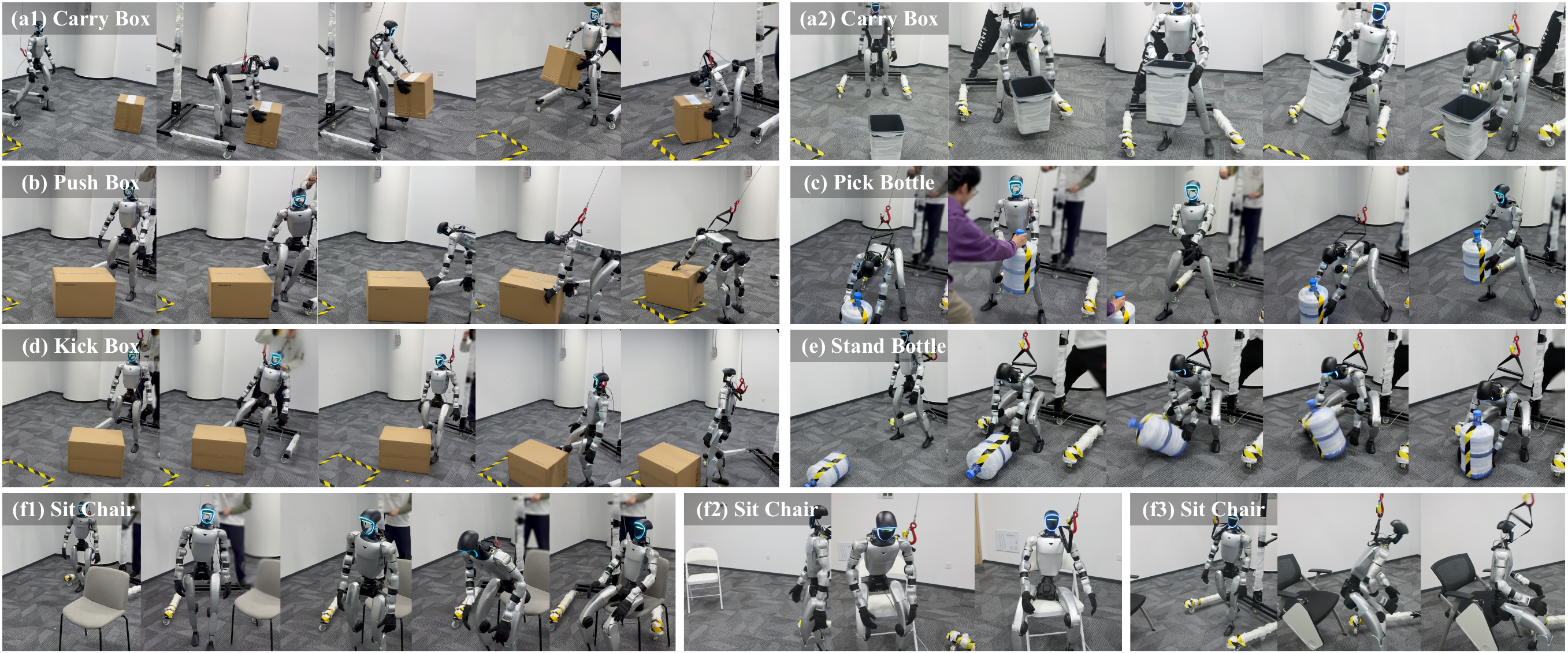}
    \caption{\textbf{\ourmethod enables generalizable real-world humanoid loco-manipulation from diverse human videos.} We deploy \ourmethod on a Unitree G1 humanoid across six representative whole-body interaction tasks: (a)~Push Box, (b)~Pick Bottle, (c)~Carry Box, (d)~Sit Chair, (e)~Kick Box, and (f1, f2)~Pick Bottle under external human disturbances.}
    \label{fig:teaser}

\end{figure}

As illustrated in Fig.~\ref{fig:method_overview}, \ourmethod proceeds in three tightly coupled stages. First, a fully automated pipeline reconstructs human motion, 6D object trajectories, and VLM-labeled contact events from unannotated videos to form scalable kinematic priors. Second, a privileged RL policy utilizes a unified mimic-style reward and a novel progressive state pool to transform
these coarse kinematic interaction priors into physically feasible and high-fidelity skills. Finally, we distill these skills into a hierarchical policy: a high-level diffusion policy command generator synthesizes movement intent commands, while a low-level whole-body command tracker robustly tracks them. 

We evaluate \ourmethod on six representative whole-body loco-manipulation tasks on a Unitree G1 humanoid. Across both training and unseen test configurations, \ourmethod substantially outperforms reference-tracking baselines, exhibits favorable scaling with the amount of human video data, and successfully deploys on real hardware with robust closed-loop execution, autonomous failure recovery, and stable long-horizon interaction under external perturbations. In summary, our contributions are:
\begin{itemize}[leftmargin=2em, itemsep=2pt]
\item A fully automated pipeline that first extracts coarse kinematic interaction priors from unstructured human videos, and subsequently refines them into physically feasible, high-fidelity skill demonstrations via privileged reinforcement learning.
\item A systematic hierarchical policy training pipeline that converts refined skill demonstrations into a deployable, reference-free autonomous policy.
\item Extensive simulation and real-world experiments validating that our method outperforms strong baselines, generalizes to unseen object configurations, scales naturally with video data, and transfers zero-shot to real hardware with robust closed-loop recovery.
\end{itemize}

%% file: tex/2_related_work.tex
\section{Related Work}

\subsection{Humanoid-Object Interaction}
\vspace{-0.5em}
Humanoid-object interaction
remains a challenging open problem. Task-specific RL in simulation produces impressive results on tasks such as soccer~\citep{wang2025learningvisiondrivenreactivesoccer, su2025realworldcooperativecompetitivesoccer}, tennis~\citep{zhang2026learningathletichumanoidtennis}, and door opening~\citep{chen2024eqvaffordse3equivariancepointlevel,xue2025openingsimtorealdoorhumanoid}, but requires per-task reward engineering~\citep{zhuang2026deepwholebodyparkour, yin2025visualmimic}. An alternative line of work collects robot demonstrations through teleoperation and trains autonomous policies from this data~\citep{chen2026learningpartawaredense3d, ze2025twist2scalableportableholistic, wei2026psi0openfoundationmodel, li2025cloneclosedloopwholebodyhumanoid, ben2025homiehumanoidlocomanipulationisomorphic, li2025amoadaptivemotionoptimization, jiang2025wholebodyvla}. However, it is bottlenecked by human effort and specialized hardware. In the character animation community, retargeting mocap data to humanoid robots and applying imitation-based rewards has enabled robots to acquire diverse locomotion and interaction skills~\citep{xu2026interprior, xu2025interact, xu2026intermimicuniversalwholebodycontrol, wang2023physhoiphysicsbasedimitationdynamic, tessler2024maskedmimicunifiedphysicsbasedcharacter, tevet2024closdclosingloopsimulation, wang2025skillmimic, yu2025skillmimicv2, wang2026omnixtremebreakinggeneralitybarrier,mahmood2019amassarchivemotioncapture, lee2025phumaphysicallygroundedhumanoidlocomotion}. However, deploying these methods on real-world humanoid robots faces substantial challenges~\citep{nai2026humi, weng2025hdmilearninginteractivehumanoid,wang2026humanxagilegeneralizablehumanoid, yang2025omniretargetinteractionpreservingdatageneration, lin2026lessmimic, he2026ultra,wang2025physhsirealworldgeneralizablenatural, fu2024humanplushumanoidshadowingimitation}: kinematic differences between human and robot embodiments, maintaining physical plausibility during HOI retargeting, and the sim-to-real gap in contact-rich interactions all hinder direct transfer. In this work, \ourmethod addresses these challenges by proposing a scalable pipeline for constructing physically grounded HOI data from human videos, combined with a hierarchical policy framework that enables generalizable object interaction on real humanoid hardware.
\vspace{-0.25em}
\subsection{Humanoid Learning from Human Videos}
\vspace{-0.5em}
Recent advances in humanoid robot learning have increasingly turned to human videos as a scalable source of demonstration data. One line of work learns locomotion skills from video demonstrations~\citep{he2025asapaligningsimulationrealworld, mao2024learningmassivehumanvideos, allshire2025visualimitationenablescontextual, yang2026zerowbclearningnaturalvisuomotor, xie2025kungfubot, han2025kungfubot2}, successfully transferring walking, running, and acrobatic behaviors to humanoid robots. However, these approaches fundamentally lack object interaction capabilities, as they do not explicitly model object dynamics during training. Another line of work focuses on learning manipulation skills from video~\citep{shi2026egohumanoidunlockinginthewildlocomanipulation, gao2026dreamdojogeneralistrobotworld, lepert2025masqueradelearninginthewildhuman, shah2025mimicdroidincontextlearninghumanoid,li2024okamiteachinghumanoidrobots, zhu2025visionbasedmanipulationsinglehuman, heng2026humdexhumanoiddexterousmanipulation}, but is typically constrained to upper-body or tabletop interactions, failing to exploit the large workspace achievable through whole-body coordination.
Recent works~\citep{weng2025hdmilearninginteractivehumanoid, zhao2025resmimicgeneralmotiontracking} take a step toward unifying locomotion and manipulation by learning whole-body interactions from monocular RGB videos, co-tracking human and object trajectories. Nevertheless, it remains a reference-based approach that replays recorded motions at inference, limiting generalization to novel objects and configurations.
Concurrent work HumanX~\citep{wang2026humanxagilegeneralizablehumanoid} compiles human video into real-world interaction skills, but relies on kinematic motion synthesis with manually defined anchor points rather than learning from large-scale multi-trajectory HOI data. In contrast, our approach automatically extracts and refines HOI data from diverse human videos at scale, and learns generalizable, reference-free interaction policies through a hierarchical architecture.

%% file: tex/3_method.tex
\section{Method}

\begin{figure}[h]
    \centering
    \includegraphics[width=1.0\textwidth]{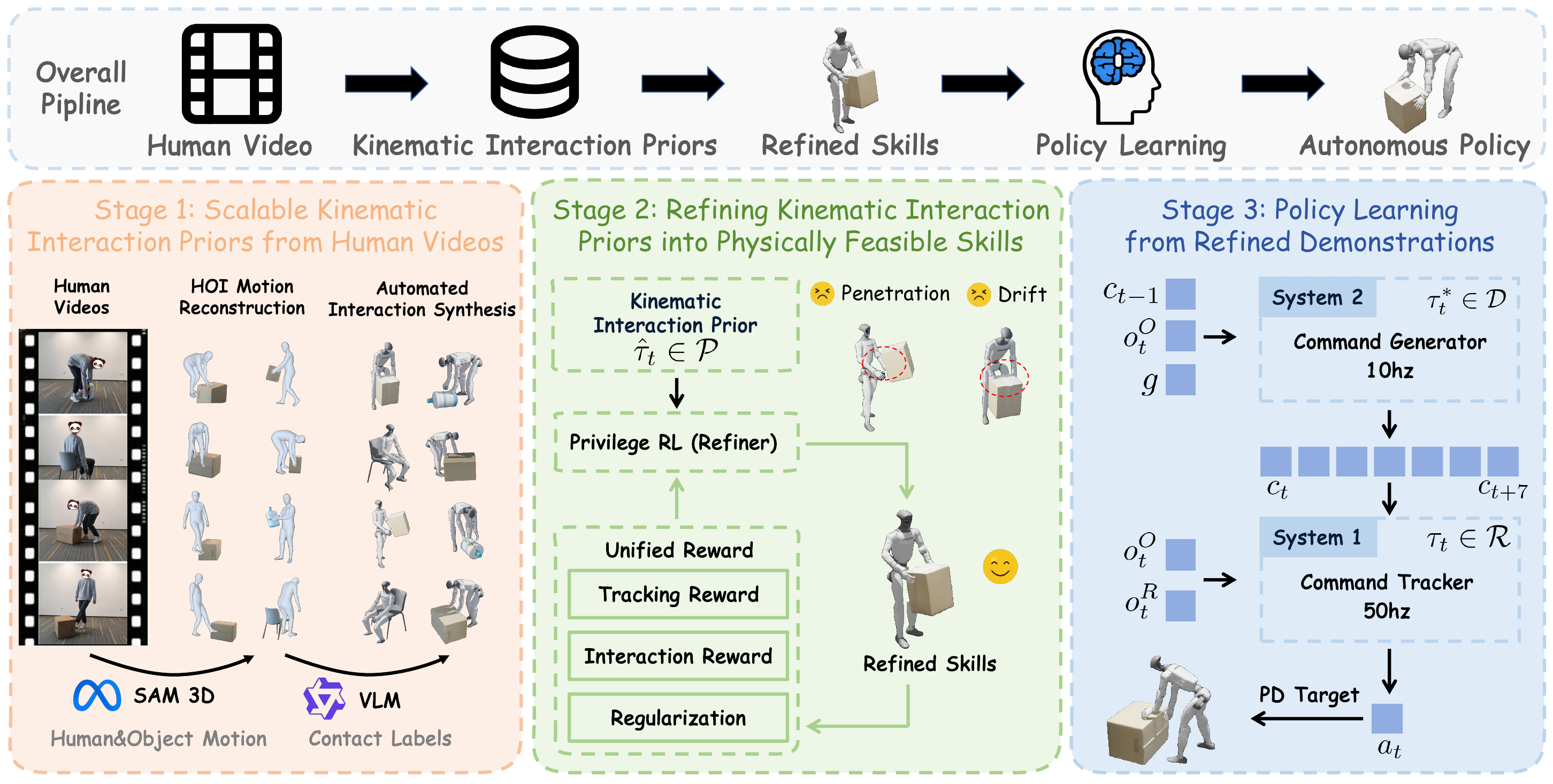}
    \vspace{-1.8em}
    \caption{\textbf{Overview of \textsc{Sugar}.} Our approach consists of three stages: (1) extracting kinematic interaction priors from unstructured human videos through a fully automated pipeline; (2) refining the priors into physically feasible skills with a privileged RL policy; and (3) training a hierarchical autonomous policy on the refined demonstrations for robust humanoid locomanipulation.}
    \label{fig:method_overview}
    \vspace{-0.4em}
\end{figure}

\vspace{-0.25em}
\subsection{Overview}
\vspace{-0.25em}
We aim to learn an autonomous policy $\pi$ for humanoid locomanipulation tasks by leveraging human videos as a primary data source.
Formally, given the robot proprioception $o_t^R$, object observation $o_t^O$ (represented as the 6D pose relative to the robot's root frame), and an optional task goal $g$ (represented as the target object state), the policy $\pi$ predicts the action $a_t$ to achieve the task: $a_t = \pi(o_t^R, o_t^O, g)$, which is then transformed into joint torques via a PD controller.

As illustrated in Fig.~\ref{fig:method_overview}, our approach consists of three core stages designed to bridge the gap between unannotated human videos and robust physical execution:
(1) We propose a fully automated pipeline to extract human-object kinematic interaction priors, including motion trajectories and contact labels, from unstructured human videos.(Sec.~\ref{sec:3_2})
(2) We train a privileged RL policy, refiner, to transform these coarse kinematic interaction priors into physically feasible and high-fidelity skills.(Sec.~\ref{sec:3_3})
(3) We learn an autonomous policy from the refined demonstrations, distilling the expert knowledge into a robust system capable of both high-level task planning and low-level command tracking.(Sec.~\ref{sec:3_4})

\subsection{Scalable Kinematic Interaction Priors from Human Videos}
\label{sec:3_2}

\vspace{-0.25em}

Human videos offer an abundant, low-cost data source for skill acquisition. However, existing methods face constraints in fully exploiting this potential. 
To address this, we propose a fully automated pipeline to extract kinematic interaction priors dataset $\mathcal{P}$ consisting of trajectories and contact labels from raw video, eliminating manual annotation labor.

\vspace{-0.4em}

\paragraph{Human-Object Motion Reconstruction}
We utilize SAMBody~\citep{yang2026sam3dbody, gao2025sambody4d} to extract human motion sequences $\hat{p}_{1:T}^R$, which are further aligned with the depth observations and optimized using Iterative Closest Point (ICP)~\citep{besl1992method} to ensure spatial accuracy.
For objects, we first generate a mesh using SAMObj~\citep{sam3dteam2025sam3d3dfyimages} and determine its physical scale by aligning the mesh with the captured object point cloud. We then employ FoundationPose~\citep{foundationposewen2024} to estimate the object's 6D pose trajectories $\hat{p}_{1:T}^O$.

\vspace{-0.4em}

\paragraph{Automated Interaction Synthesis}
To bypass manual annotation when assigning contact labels $\hat{l}_t$, we query a VLM~\citep{Qwen3-VL} based on task-specific body parts (e.g., hands for carrying), the prompt will be shown in Appendix~\ref{contact_label_genetation}.
In tasks where visual cues are ambiguous due to severe occlusion (e.g., kicking box), preventing the VLM from providing reliable per-frame contact signals, we infer contact if the object's velocity exceeds the threshold.

Finally, we apply temporal filtering to smooth the trajectories, yielding a comprehensive dataset of kinematic interaction priors dataset $\mathcal{P}$, which serves as a structured reference for subsequent physics-based refinement:
\begin{equation}
\mathcal{P} = \{ \hat\tau_i \}_{i=1}^N, \quad \text{where } \hat\tau = \{ (\hat{p}_t^R, \hat{p}_t^O, \hat{l}_t) \}_{t=1}^T
\label{eq:dataset_p}
\end{equation}
Each trajectory $\hat{\tau}$ represents a sequence of reconstructed human-object motions and contact labels extracted from a single video clip. The "hat" notation ($\hat{\cdot}$) signifies that these priors are derived from purely kinematic estimation and may contain physical inaccuracies.

\begin{figure}[t]
    \centering
     \includegraphics[width=1.0\textwidth]{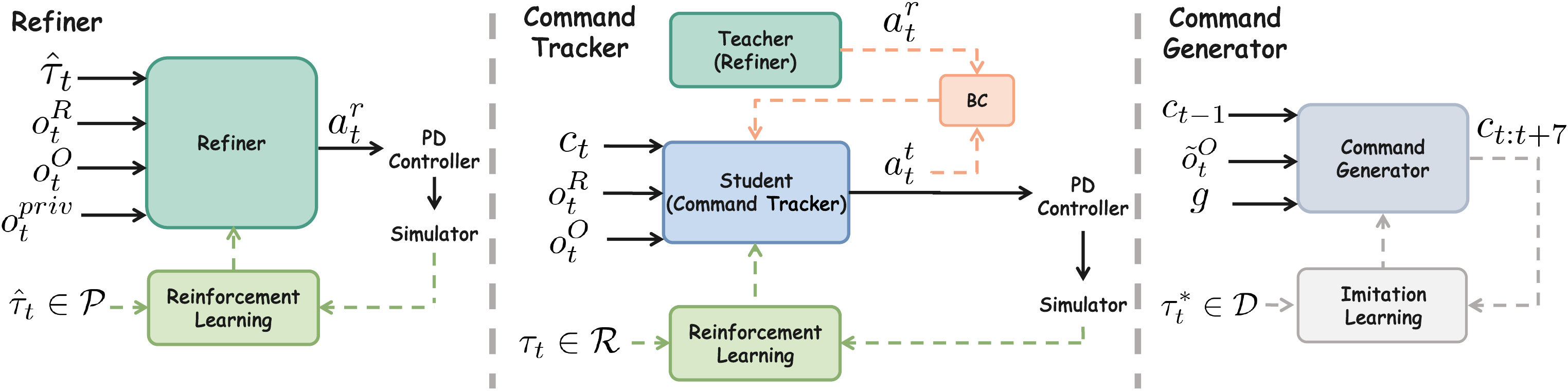}
     \vspace{-1.5em}
    \caption{The Training Pipeline of \textsc{Sugar}.
    (Left) The Refiner $\pi_r$ transforms noisy kinematic priors $\hat{\tau} \in \mathcal{P}$ into physically feasible expert demonstrations $\tau \in \mathcal{R}$ using privileged RL.
    (Middle) The Tracker $\pi_t$ distills motor skills from the Refiner via behavior cloning and reinforcement learning to achieve robust command-tracking.
    (Right) The Generator $\pi_g$ is trained via imitation learning on the rollout dataset $\mathcal{D}$ to predict high-level command sequences, enabling autonomous locomanipulation in a hierarchical manner.}
    \label{fig:training_pipeline}
    \vspace{-0.7em}
\end{figure}

\subsection{Refining Kinematic Interaction Priors into Physically Feasible Skills}
\label{sec:3_3}

\vspace{-0.25em}

To bridge the gap between kinematic priors and physical dynamics, we train a privileged reference-tracking RL policy, refiner, $\pi_{r}(a_t^r \mid o_t^R, o_t^O, o_t^{priv}, \hat{\tau}_i)$ that translates noisy $\mathcal{P}$ into physically feasible expert trajectories while maintaining the original task intent, producing a physically feasible refined skill dataset $\mathcal{R}$:
\begin{equation}
\mathcal{R} = \{ \tau_i \}_{i=1}^N, \quad \text{where } \tau = \{ (p_t^R, p_t^O, l_t, c_t) \}_{t=1}^T
\label{eq:dataset_r}
\end{equation}
Here, $p_t^R$, $p_t^O$, and $l_t$ represent the physically consistent robot poses, object poses, and actual contact states, respectively.
To provide a concrete learning target for the subsequent stage, dataset $\mathcal{R}$ also includes the expert states ${c}_t$ recorded during successful executions, defined as:
\begin{equation}
{c}_t = [{q}^{\text{cmd}}_t, {v}_{\text{t}}^{\text{cmd}}, {\omega}_{\text{t}}^{\text{cmd}}, l_t]
\label{eq:command_vector}
\end{equation}
Specifically, ${q}_t^\text{cmd}$ denotes the actual joint positions executed by the refiner, while ${v}_{\text{t}}^\text{cmd}$ and ${\omega}_{\text{t}}^\text{cmd}$ are the resulting root linear and angular velocities. 
These recorded states serve as the reference commands for training the autonomous policy in the next stage.

\vspace{-0.5em}

\paragraph{Unified Reward Design}
To minimize task-specific engineering, we design a unified mimic reward $r = r_{track} + r_{int} + r_{reg}$, with detailed formulations provided in Appendix~\ref{reward_details}.
The \textbf{tracking reward term} $r_{track}$ encourages the robot and object to follow reference trajectories $(\hat{p}_t^R, \hat{p}_t^O)$. Notably, the robot tracks root-relative poses to isolate reconstruction errors and prevent noisy global drift from degrading motion naturalness.
The \textbf{interaction reward term} $r_{int}$ enforces physical consistency by utilizing contact labels $\hat{l}_t$ to prevent force-inconsistent motions and penalizing spatial decoupling between robot links and the object.
Finally, \textbf{regularization term} $r_{reg}$ facilitates sim-to-real transfer by penalizing torque and non-smooth behaviors.

\vspace{-0.5em}

\paragraph{Progressive State Pool for Initialization}
Standard Reference State Initialization (RSI) often fails in HOI tasks due to kinematic reconstruction errors, such as penetrations and misalignments, which create physically infeasible starting points. To mitigate this, we propose the Progressive State Pool $\mathcal{B}$, which initializes agents from physically-validated states instead of unreliable references from $\mathcal{P}$. During training, $\mathcal{B}$ is incrementally populated with successful intermediate states encountered by $\pi_r$, providing diverse, physically consistent milestones that stabilize learning of complex interactions while preventing overfitting.

\vspace{-0.5em}

\paragraph{Interaction Robustness Enhancement}
We apply extensive randomizations and perturbations to broaden the state-dynamics coverage.
By varying physical properties like mass and friction, and applying random impulses to both the robot and the object, we force the refiner to learn real physical rules instead of finding shortcuts in the simulator.
This ensures the policy develops robust interaction habits that work under uncertain conditions. Consequently, the generated expert data is physically sound and stays stable under interference, providing a reliable basis for training.

\subsection{Policy Learning from Refined Demonstrations}
\label{sec:3_4}

\vspace{-0.25em}

Given the refined skill dataset $\mathcal{R}$, we aim to train a policy capable of autonomous real-world execution.
We separate the autonomous policy into two functional components: a Command Generator for task-level intent and a Command Tracker for robust physical execution.

\vspace{-0.5em}

\paragraph{Command Tracker}
The tracker $\pi_{t}(a_t^t \mid o_t^R, o_t^O, {c}_t)$ is designed to follow a movement intent command ${c}_t$ by predicting joint targets $a_t$.
During the distillation phase, ${c}_t$ is from the refined dataset $\mathcal{R}$ as an expert reference, whereas during autonomous inference, it is actively produced by the command generator $\pi_g$, as illustrated in Fig.~\ref{fig:training_pipeline}.
These targets are subsequently converted into joint torques via a PD controller.

\vspace{-0.5em}

\begin{itemize}[leftmargin=2em, itemsep=2pt]
    \item \textbf{Distillation from Refiner.}
    To efficiently scale $\pi_{t}$'s capability across diverse demonstrations, we distill the expertise from the refiner $\pi_{r}$ combining Behavior Cloning (BC) and Reinforcement Learning (RL).
    Specifically, $\pi_{t}$ first performs BC to rapidly mimic $\pi_{r}$’s motion patterns, providing a structured initialization.
    This is followed by a transitional phase to warm up the critic and actor of $\pi_{t}$, eventually shifting to full RL optimization using the same rewards as in Sec.~\ref{sec:3_3}.
    \item \textbf{Evolutionary Initialization.}
    Leveraging the physically consistent states provided by the $\mathcal{R}$, we first utilize Reference State Initialization (RSI) to ensure a stable start.
    As training progresses, we shift to sampling from the Progressive State Pool Initialization (PSPI), to broaden the state coverage and prevent over-fitting to single demonstrations. This allows $\pi_{t}$ to reliably master complex, multi-stage interactions from a stable physical foundation.
\end{itemize}

\vspace{-1.0em}

\paragraph{Task-Guided Command Generator}With the low-level tracker $\pi_{t}$ frozen, autonomous locomanipulation is reformulated as a conditional sequence generation problem. We implement the high-level task-guided command generator $\pi_{g}
(c_{t:t+7} \mid o_t^O, c_{t-1}, g)$ using a state-based Diffusion Policy~\citep{chi2025diffusion}, which predicts a sequence of future commands to drive the tracker toward task completion. We defer more details into Appendix~\ref{appendix_dp}.

To bridge the gap between kinematic planning and dynamic execution, we collect a rollout dataset $\mathcal{D}$ that reflects the actual performance of the integrated system.
Specifically, for each refined trajectory $\tau \in \mathcal{R}$, we drive the frozen tracker $\pi_t$ using the recorded expert states $\mathbf{c}_t$ as reference commands. We then record the actual object states $\tilde{o}_t^O$ reached by the tracker during these closed-loop rollouts, forming the dataset:
\begin{equation}
\mathcal{D} = \{ \tau_i^* \}_{i=1}^M, \quad \text{where } \tau_i^* = \{ (\tilde{o}_t^O, {c}_t, g) \}_{t=1}^T
\label{eq:dataset_d}
\end{equation}

By training on these execution-based data rather than idealized references, the command generator learns to guide the tracker based on the states it encounters. This approach enables the generator to proactively correct execution errors and drift, ensuring reliable task completion over long horizons.

\vspace{-0.5em}

%% file: tex/4_experiment.tex
\section{Experiment}
\vspace{-0.4em}
To evaluate the effectiveness of our method, we design a series of experiments to answer the following questions:
(1) Performance and Generalization: How does our method perform compared to prior approaches in terms of performance and generalization to unseen initial and target states?
(2) Data Scaling: How does the performance of our method improve as the amount of training data increases?
(3) Component Analysis: How does each component contribute to our framework?
(4) Sim-to-Real Transfer: Can the learned policy be robustly transferred from simulation to real-world deployment?

\subsection{Experiment Setup}

\vspace{-0.25em}

\paragraph{Tasks.}
We evaluate our method on six challenging whole-body loco-manipulation tasks:
(1) \textit{Carry Box}: lift and transport a box to the target location;
(2) \textit{Push Box}: push a box to the target location;
(3) \textit{Kick Box}: kick a box to the target location;
(4) \textit{Pick Bottle}: walk to and lift a bottle from the ground;
(5) \textit{Stand Bottle}: reorient a bottle from lying to upright pose;
(6) \textit{Sit Chair}: move from varied initial positions and stably sit on a chair.

\vspace{-0.4em}

\paragraph{Dataset.}
For each task, we collect 100 human video demonstrations for training and 30 for testing. By leveraging human videos rather than robot teleoperation, data collection remains efficient and low-cost. Our pipeline then automatically processes these videos into training-ready data.

\vspace{-0.4em}

\paragraph{Evaluation Metrics.}
We use the following metrics:
(1) \textit{Success Rate}: the percentage of successful trials.
A trial is considered successful based on task-specific criteria. 
For \textit{Carry Box}, \textit{Push Box}, and \textit{Kick Box}, success is defined as the final object position being within a predefined threshold of the target location. 
For \textit{Sit Chair}, success requires the robot base to maintain stable contact with the chair for a certain duration. 
For \textit{Stand Bottle}, success is achieved when the bottle is stably placed in an upright position on the ground. 
For \textit{Pick Bottle}, success is defined as lifting the bottle above a predefined height.
(2) \textit{Final Object Position Error}: the Euclidean distance between the final object position and the target location for tasks involving target placement.

\vspace{-0.5em}

\paragraph{Baselines.}
We compare our method with two representative methods, \textit{Resmimic}~\citep{zhao2025resmimicgeneralmotiontracking} and \textit{HDMI}~\citep{weng2025hdmilearninginteractivehumanoid}.
These baselines represent strong prior methods based on reference trajectory replay. All methods are trained and evaluated under the same dataset.

\input{tables/main}

\vspace{-0.5em}
\subsection{Comparison with Baselines}
\vspace{-0.3em}
We compare our method with baseline methods on six whole-body loco-manipulation tasks. 
Table~\ref{tab:main_results} shows our method outperforms the baselines on all tasks in both success rate and final object position error.
The performance gap is most evident in high-precision tasks like \textit{Carry Box}, \textit{Pick Bottle}, and \textit{Stand Bottle}. While baseline methods fail to learn effective skills from coarse, noisy dataset, our approach consistently extracts reusable skills and achieves significantly higher success rates.

\vspace{-0.5em}
\subsection{Performance scaling with data size}
\begin{figure}[t] 
    \centering    
\includegraphics[width=1.0\textwidth]{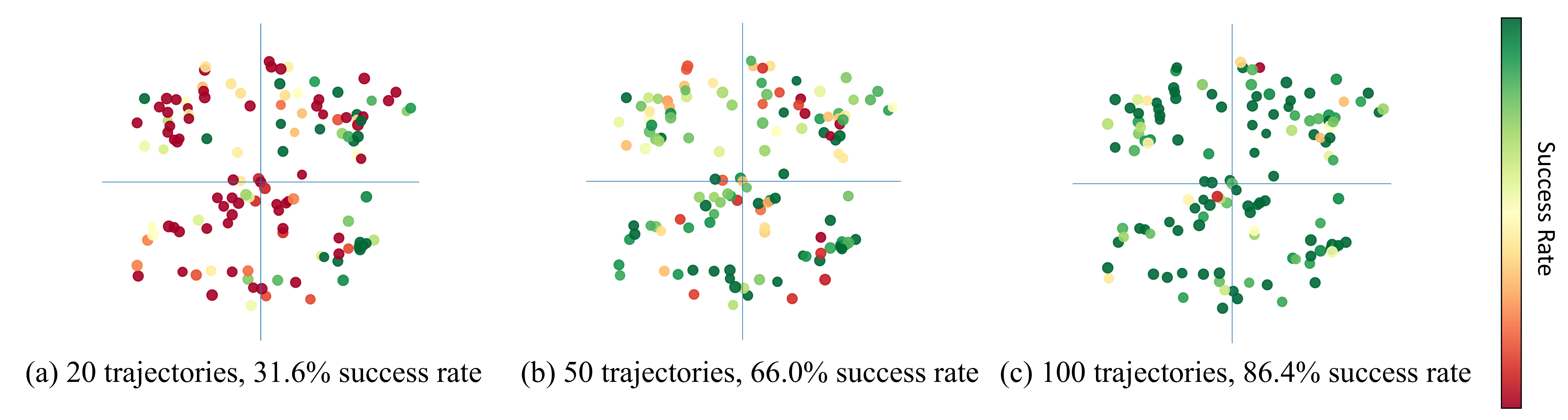} 
    \caption{\textbf{Performance with different training data sizes.} Success rates, evaluated on both the train
and test datasets, consistently improve as the amount of training data increases.} 
    \label{fig:datascale}  
\end{figure}

\vspace{-0.3em}
To analyze performance scaling, we train our model using 20, 50, and 100 trajectories per task.
As shown in Table~\ref{tab:datascale_result} and Fig~\ref{fig:datascale}, our method exhibits a strong scaling trend, where success rates improve consistently as the data volume increases.
The improvement with more data is primarily due to increased coverage of state-action space.
This suggests that our architecture can inherently capture more robust and generalizable behaviors from larger datasets without requiring additional task-specific rewards or robustness engineering. Such scalability highlights the potential of our approach to benefit from large-scale noisy human data in complex scenarios.

\input{tables/datascale}

\vspace{-0.5em}
\subsection{Component Analysis}
\vspace{-0.25em}
We conduct ablation studies to evaluate the contribution of key components in our framework.

\textbf{Refinement Policy (w/o Refiner):} Removing the Refiner leads to substantial performance degradation, as shown in Table~\ref{tab:main_results}, proving that physical consistency is crucial when learning from noisy video data.
This result highlights that the Refiner transforms coarse motion priors into physically valid and dynamically consistent demonstrations, enables stable and effective learning.

\textbf{Progressive State Pool (w/o PSPI):} We analyze the impact of the Progressive State Pool for Initialization by replacing it with two alternatives:
(1) \textit{Start State Initialization (SSI)}, where training always initializes from start phase; and 
(2) \textit{Reference State Initialization (RSI)}, where initial states are sampled from raw kinematic trajectories.
Both variants lead to noticeable performance degradation (Table~\ref{tab:main_results}).
In contrast, the Progressive State Pool provides diverse and physically consistent initialization states, enabling stable training and effective skill acquisition across different stages.

\begin{figure}[t] 
    \centering    
\includegraphics[width=1.0\textwidth]{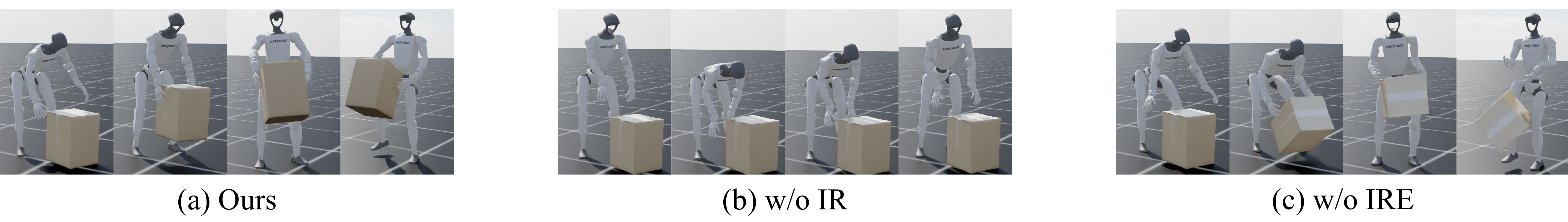} 
    \caption{\textbf{Qualitative results: \textit{Carry Box}.} (a) Our method stably lifts the box. (b) Without interaction rewards (w/o IR), the policy only imitates the bending motion and fails to lift the box (c) Without interaction robustness enhancement (w/o IRE), the interaction is less robust and causes failure.} 
    \label{fig:componentAnalysis}  
\end{figure}

\input{tables/realworld}

\begin{figure}[t]
    \centering 
    \includegraphics[width=1.0\textwidth]{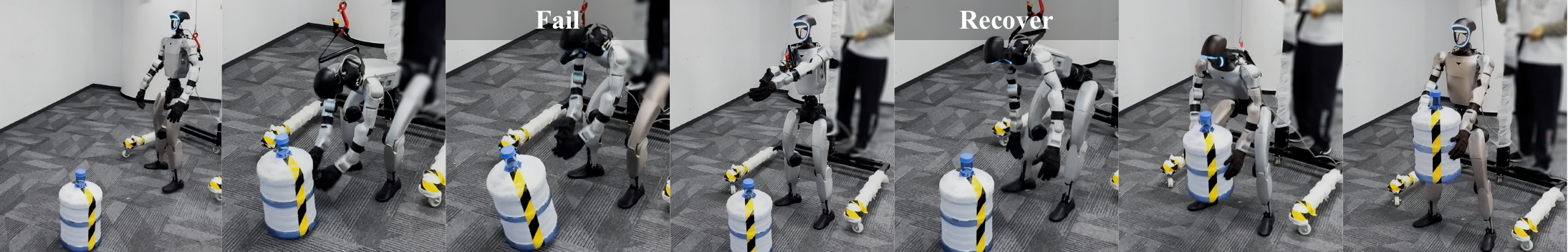}
    \vspace{-1.8em}
    \caption{\textbf{Recover from failure.}} 
    \label{fig:recovery} 
    \vspace{-0.2em}
\end{figure}

\textbf{Interaction Rewards (w/o IR):} Quantitative and qualitative results (Table.~\ref{tab:main_results} and Fig.~\ref{fig:componentAnalysis}) show removing interaction rewards leads to failure in contact-rich tasks like \textit{Carry Box} and \textit{Pick Bottle}, as kinematic tracking alone fails to enforce essential physical constraints.

\textbf{Interaction Robustness Enhancement (w/o IRE):}
As also illustrated in Table.~\ref{tab:main_results} and Fig.~\ref{fig:componentAnalysis}, removing Interaction Robustness Enhancement causes severe overfitting to idealized physics, incorporating it forces the model to learn physical compensation, significantly improving robustness against external disturbances and varying physical properties.

\vspace{-0.2em}

\subsection{Real-World Evaluation}
\vspace{-0.15em}
\begin{figure}[h!] 
    \centering  
    \includegraphics[width=1.0\textwidth]{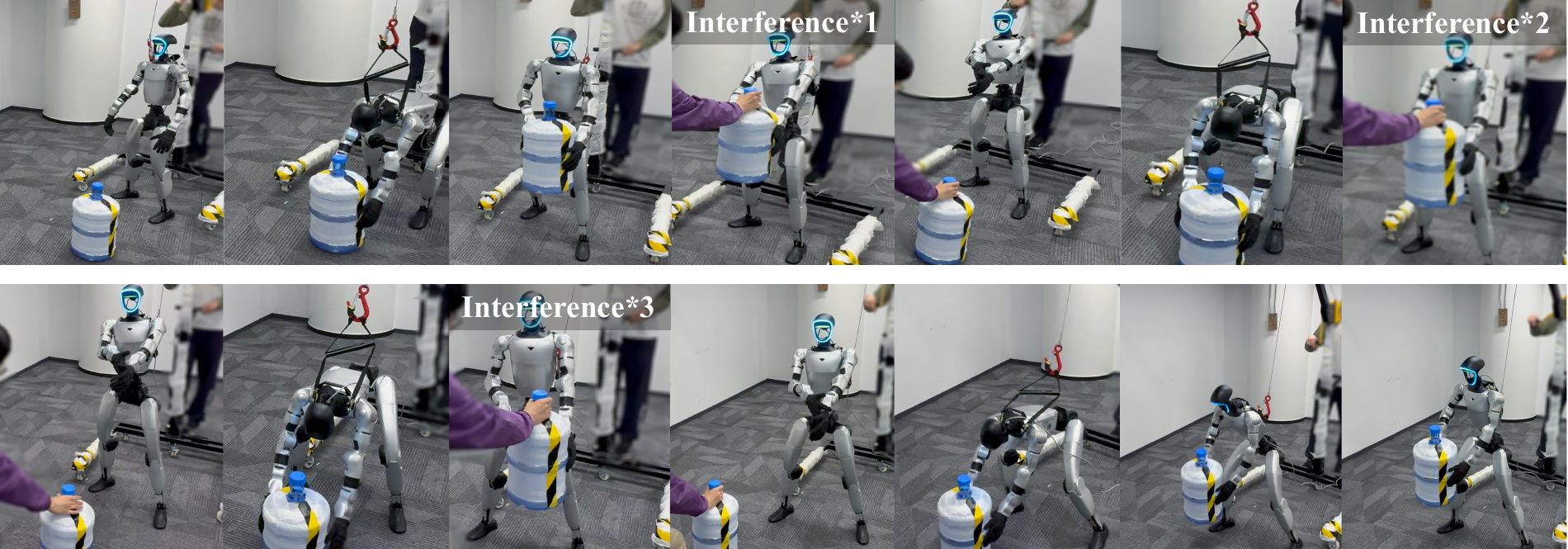} 
    \vspace{-1.7em}
    \caption{\textbf{Robustness to external disturbances in the real world.}} 
    \label{fig:interference}  
    \vspace{-0.6em}
\end{figure}

\begin{figure}[h!] 
    \centering  
    \includegraphics[width=1.0\textwidth]{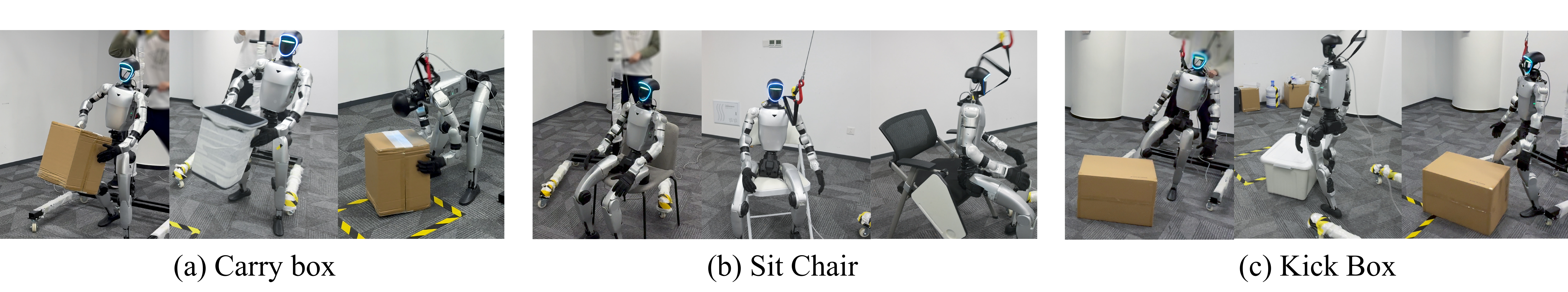} 
    \vspace{-1.9em}
    \caption{\textbf{Zero-shot generalization to different objects in the real world.}} 
    \label{fig:realworldGeneralization} 
    \vspace{-0.3em}
\end{figure}

We deploy our policy on a real humanoid robot using MoCap, transferring the purely simulation-trained model to the real world.
We evaluate each task over 10 trials and summarize the success rates in Table~\ref{tab:realworld_sr}.
In real-world experiments, the learned policy demonstrates robust closed-loop execution under noisy perception and dynamical discrepancies, enabling the robot to continuously perform tasks over extended horizons while maintaining consistent task progress.

A key observation is the policy’s robustness in real-world execution. As shown in Fig.~\ref{fig:recovery}, when execution is disrupted, such as by object displacement or partial task failure, the robot can autonomously resume the task rather than terminating, indicating the ability to handle out-of-distribution states.
As illustrated in Fig.~\ref{fig:interference}, the robot also remains stable under external disturbances and continues execution without losing control. Moreover, Fig.~\ref{fig:realworldGeneralization} shows that the learned policy generalizes zero-shot to objects with different shapes, sizes, and appearances without finetuning, suggesting that it captures transferable interaction strategies instead of overfitting to a specific object instance.

%% file: tables/main.tex
\begin{table*}[t]
\centering
\caption{\textbf{Main results in simulation.} We evaluate two baselines, ablated variants, and \ourmethod on six whole-body loco-manipulation tasks in simulation. The two baselines additionally require reference demonstration trajectory observations, whereas \ourmethod takes only an optional goal object state $g$ as input . All methods are trained and evaluated on the same training and test datasets.}
\vspace{0.5em}
\label{tab:main_results}

\small

\begin{tabularx}{\textwidth}{lZZZZZZ}
\toprule

\multirow{2}{*}{\textbf{Method}}
& \textbf{\mbox{Kick Box}}
& \textbf{\mbox{Push Box}}
& \textbf{\mbox{Carry Box}}
& \textbf{\mbox{Sit Chair}}
& \textbf{\mbox{Pick Bottle}}
& \textbf{\mbox{Stand Bottle}} \\

\cmidrule(lr){2-7}

& \twometrics{SR$\uparrow$}{Err$\downarrow$}
& \twometrics{SR$\uparrow$}{Err$\downarrow$}
& \twometrics{SR$\uparrow$}{Err$\downarrow$}
& SR$\uparrow$
& SR$\uparrow$
& SR$\uparrow$ \\

\midrule
\multicolumn{7}{c}{\textbf{Training Dataset}} \\
\midrule

\textbf{Resmimic}
& \twometrics{21.6}{0.912}
& \twometrics{16.5}{1.021}
& \twometrics{0.0}{$-$}
& 23.2
& 0.0
& 0.0 \\

\textbf{HDMI}
& \twometrics{82.4}{0.263}
& \twometrics{\textbf{94.1}}{\textbf{0.162}}
& \twometrics{0.0}{$-$}
& 31.9
& 0.0
& 0.0 \\

w/o refiner
& \twometrics{48.3}{0.419}
& \twometrics{56.5}{0.464}
& \twometrics{66.8}{0.429}
& 93.4
& 97.5
& 79.0 \\

w/o PSPI (SSI)
& \twometrics{80.9}{0.251}
& \twometrics{78.5}{0.272}
& \twometrics{81.3}{0.303}
& 95.0
& 98.2
& 89.6 \\

w/o PSPI (RSI)
& \twometrics{85.9}{0.241}
& \twometrics{83.9}{0.272}
& \twometrics{68.9}{0.483}
& \textbf{96.7}
& 94.3
& 86.2 \\

w/o IR
& \twometrics{80.4}{0.254}
& \twometrics{85.5}{0.264}
& \twometrics{73.8}{0.390}
& 96.0
& 0.0
& 85.0 \\

w/o IRE
& \twometrics{86.4}{0.237}
& \twometrics{72.4}{0.328}
& \twometrics{67.7}{0.388}
& 92.8
& 97.0
& 59.4 \\

\textbf{Ours}
& \twometrics{\textbf{89.5}}{\textbf{0.222}}
& \twometrics{83.6}{0.261}
& \twometrics{\textbf{84.5}}{\textbf{0.280}}
& 94.8
& \textbf{98.8}
& \textbf{91.9} \\

\midrule
\multicolumn{7}{c}{\textbf{Test Dataset}} \\
\midrule

\textbf{Resmimic}
& \twometrics{18.5}{0.963}
& \twometrics{10.6}{1.160}
& \twometrics{0.0}{$-$}
& 20.6
& 0.0
& 0.0 \\

\textbf{HDMI}
& \twometrics{17.3}{0.938}
& \twometrics{54.6}{0.377}
& \twometrics{0.0}{$-$}
& 8.2
& 0.0
& 0.0 \\

w/o refiner
& \twometrics{46.3}{0.414}
& \twometrics{41.7}{0.554}
& \twometrics{62.0}{0.424}
& 99.0
& 96.7
& 79.0 \\

w/o PSPI (SSI)
& \twometrics{61.7}{0.316}
& \twometrics{71.0}{0.328}
& \twometrics{63.3}{0.337}
& 98.0
& 94.9
& 84.3 \\

w/o PSPI (RSI)
& \twometrics{71.4}{0.269}
& \twometrics{\textbf{73.0}}{0.345}
& \twometrics{62.0}{0.557}
& 98.3
& 94.2
& 79.6 \\

w/o IR
& \twometrics{\textbf{76.3}}{\textbf{0.260}}
& \twometrics{63.4}{0.338}
& \twometrics{60.3}{0.513}
& 97.6
& 0.0
& 73.1 \\

w/o IRE
& \twometrics{72.0}{0.298}
& \twometrics{47.2}{0.421}
& \twometrics{57.5}{0.436}
& 98.9
& 98.0
& 63.4 \\

\textbf{Ours}
& \twometrics{76.0}{0.265}
& \twometrics{70.0}{\textbf{0.325}}
& \twometrics{\textbf{69.6}}{\textbf{0.326}}
& \textbf{99.6}
& \textbf{99.2}
& \textbf{86.3} \\

\bottomrule
\end{tabularx}
\vspace{-1.1em}
\end{table*}

%% file: tables/datascale.tex
\begin{table*}[t]
\centering
\caption{\textbf{Simulation results under different training data sizes.} We evaluate \ourmethod on six whole-body loco-manipulation tasks using 20, 50, and 100 training trajectories per task. Success rates improve consistently as the amount of training data increases.}
\vspace{0.6em}
\label{tab:datascale_result}

\small

\begin{tabularx}{\textwidth}{lZZZZZZ}
\toprule

\multirow{2}{*}{\textbf{Method}}
& \textbf{\mbox{Kick Box}}
& \textbf{\mbox{Push Box}}
& \textbf{\mbox{Carry Box}}
& \textbf{\mbox{Sit Chair}}
& \textbf{\mbox{Pick Bottle}}
& \textbf{\mbox{Stand Bottle}} \\

\cmidrule(lr){2-7}

& \twometrics{SR$\uparrow$}{Err$\downarrow$}
& \twometrics{SR$\uparrow$}{Err$\downarrow$}
& \twometrics{SR$\uparrow$}{Err$\downarrow$}
& SR$\uparrow$
& SR$\uparrow$
& SR$\uparrow$ \\

\midrule
\multicolumn{7}{c}{\textbf{Training Dataset}} \\
\midrule

20
& \twometrics{31.3}{0.518}
& \twometrics{37.1}{0.634}
& \twometrics{38.5}{0.675}
& 74.9
& 95.6
& 76.4 \\

50
& \twometrics{66.9}{0.330}
& \twometrics{30.1}{0.504}
& \twometrics{72.3}{0.367}
& 89.6
& 97.7
& 78.9 \\

100
& \twometrics{\textbf{89.5}}{\textbf{0.222}}
& \twometrics{\textbf{83.6}}{\textbf{0.261}}
& \twometrics{\textbf{84.5}}{\textbf{0.280}}
& \textbf{94.8}
& \textbf{98.8}
& \textbf{91.9} \\

\midrule
\multicolumn{7}{c}{\textbf{Test Dataset}} \\
\midrule

20
& \twometrics{32.7}{0.416}
& \twometrics{35.0}{0.593}
& \twometrics{33.5}{0.682}
& 90.0
& 94.2
& 66.4 \\

50
& \twometrics{63.1}{0.336}
& \twometrics{52.3}{0.527}
& \twometrics{61.0}{0.403}
& 98.9
& 95.3
& 75.3 \\

100
& \twometrics{\textbf{76.0}}{\textbf{0.265}}
& \twometrics{\textbf{70.0}}{\textbf{0.325}}
& \twometrics{\textbf{69.6}}{\textbf{0.326}}
& \textbf{99.6}
& \textbf{99.2}
& \textbf{86.3} \\

\bottomrule
\end{tabularx}
\end{table*}

%% file: tables/realworld.tex
\begin{table*}[t]
\centering
\caption{\textbf{Real-world success rates.} We evaluate \ourmethod on six whole-body loco-manipulation tasks in the real world. Success rates are reported as successful attempts out of 10 trials.}
\vspace{0.45em}
\label{tab:realworld_sr}

\small
\setlength{\tabcolsep}{6.5pt}

\begin{tabular}{lcccccc}
\toprule
\textbf{Kick Box} & \textbf{Push Box} & \textbf{Carry Box} & \textbf{Sit Chair} & \textbf{Pick Bottle} & \textbf{Stand Bottle} \\
\midrule
7/10 & 6/10 & 7/10 & 9/10 & 9/10 & 8/10 \\
\bottomrule
\end{tabular}
\vspace{0.5em}
\end{table*}

%% file: tex/5_conclusion.tex
\section{Conclusion}
In this work, we introduced \textsc{Sugar}, a data-driven learning framework that successfully unlocks generalizable humanoid loco-manipulation skills from diverse, unconstrained human videos. To address the physical implausibility and artifacts inherent in video-derived motion priors, we implement a robust three-stage pipeline: automated kinematic interaction prior extraction, privileged physics-based refinement via a unified reward, and hierarchical policy distillation. 
Our extensive evaluations on the Unitree G1 demonstrate that \ourmethod scales robustly with data volume,  successfully transfers to the real world, and maintains successful task completion under external disturbances, providing a scalable path for learning from human videos.

%% file: tex/6_appendices.tex
\newpage
\section{Algorithm Design}
\subsection{Implementation of RL-based Policies: Refiner and Tracker}

\paragraph{PPO Hyperparameters}
Both the Refiner ($\pi_r$) and the Command Tracker ($\pi_t$) are implemented as a three-layer MLP, optimized by PPO~\citep{schulman2017proximal}.
The detailed hyperparameters for the PPO algorithm, including network dimensions, learning rates, and clip parameters, are summarized in Table~\ref{tab:ppo_hyperparameters}.
\input{tables/ppo_config}

\paragraph{Observation Spaces}
We adopt an \textbf{asymmetric actor-critic} training scheme. The Refiner (both actor and critic) and the Tracker's critic have access to \textbf{privileged observations} shown in Table~\ref{tab:privileged_obs}. In contrast, the Tracker's actor only utilizes \textbf{deployable observations} shown in Table~\ref{tab:deployable_obs} to ensure a seamless sim-to-real transition. Notably, all robot body poses and object poses are expressed in the robot's root-relative coordinate frame to maintain translation invariance.

\input{tables/deployable_obs}
\input{tables/privileged_obs}

\paragraph{Reward Function}
\label{reward_details}
The reward function $r = r_{track} + r_{int} + r_{reg}$ balances imitation accuracy and physical feasibility. As detailed in Table~\ref{tab:rewards}:

\input{tables/reward}

\paragraph{Domain Randomization}
To make the policies robust against environmental uncertainty, we apply extensive Domain Randomization, as shown in Table~\ref{tab:domain_randomization}. By varying physical properties and applying random impulses to both the robot and the object, we force the policies to learn real physical rules instead of exploiting simulator shortcuts.
Notably, impulses on the object are only applied when active contact between the robot and the object is detected, ensuring the policies learn to maintain stable manipulation under dynamic disturbances.
\input{tables/domain_rand}

\paragraph{Early Termination}
We define several early termination terms shown in Table~\ref{tab:terminations} to reset the environment when the robot deviates excessively from the reference motion or the target task, preventing the policy from exploring unrecoverable states.
\input{tables/early_termination}

\subsection{Implementation of IL-based Policies: Command Generator}
\label{appendix_dp}
The Command Generator $\Phi_\theta$ is implemented as a 12-block Diffusion Transformer (DiT) to model the trajectory distribution~\citep{zhong2025dexgraspvla, chen2026learningpartawaredense3d}. The object state, previous command, and optional target pose are embedded via individual MLPs, which are then concatenated to form the global condition feature $F_{cond}$. 
Then the DiT backbone processes the noisy input $x_t$ at timestep $t$ conditioned on $F_{cond}$ to predict the noise residual $\hat{\epsilon}$.

During joint hierarchical inference, the Generator predicts a chunk of $H=8$ steps. To balance reactivity and smoothness, we only execute the first $A_a=4$ steps before re-planning. The outputs of the Generator are linearly interpolated to $50\,\text{Hz}$ to align with the control frequency of the Tracker.

\section{VLM-based Contact Detection Template}
\label{contact_label_genetation}
To ensure consistency across diverse scenarios, we use a unified prompt template for all tasks. 
The \texttt{[BODY\_PART]} and \texttt{[OBJECT]} are specified by the task definition. 
This approach allows the VLM to focus on the essential physical contact required for each task type without manual intervention.

\begin{quote}
\textit{Task: Determine whether the \texttt{[BODY\_PART]} is in DIRECT PHYSICAL CONTACT with the \texttt{[OBJECT]}. \\
Answer 'Yes' ONLY if the \texttt{[BODY\_PART]} is actually contacting with the \texttt{[OBJECT]}. \\
Answer 'No' if the \texttt{[BODY\_PART]} is moving toward but not touching, or if a visible gap exists. \\
Important: Do NOT infer based on intention; Do NOT predict future contact; Only judge only on actual physical contact. \\
Output: [Yes/No]}
\end{quote}

\clearpage
\section{Limitations and Future Work}
First, the current data-processing pipeline extracts relatively coarse priors, limiting the framework to coarse-grained interaction skills. How to acquire fine-grained skills remains an open question. Second, data utilization efficiency is relatively low. Exploring data augmentation and generative models to learn interaction skills from limited human videos represents a valuable direction. Finally, the state-based policy hinders deployment convenience. How to develop policies that can effectively process visual and language inputs remains an open challenge for future research.

\section{Computation Resources}
All simulation, RL policy training, and hierarchical policy inference are conducted within the IsaacSim on a single NVIDIA GeForce RTX 5090 GPU.  For each individual task, training the Refiner and the Tracker takes approximately 20 GPU hours each, while training the Command Generator requires around 5 GPU hours.

%% file: tables/ppo_config.tex
\begin{table}[h]
\centering
\caption{PPO Learning Hyperparameters}
\label{tab:ppo_hyperparameters}
\small
\begin{tabular}{@{}lc@{}}
\toprule
\textbf{Hyperparameter} & \textbf{Value} \\ \midrule
Actor MLP network & [512, 256, 128] \\
Critic MLP network & [512, 256, 128] \\
Activation function & ELU \\
Initial noise std (Refiner) & 1.0 \\ 
Initial noise std (Tracker) & 0.5 \\ \midrule
Training iterations & 30,000 \\
Number of envs & 4096 \\
Steps per env & 24 \\
Number of mini-batches & 4 \\
Number of learning epochs & 5 \\
Learning rate & 1e-3 \\
Desired KL divergence & 0.01 \\ \midrule
Discount factor ($\gamma$) & 0.99 \\
GAE parameter ($\lambda$) & 0.95 \\
PPO clip parameter & 0.2 \\
Entropy coefficient & 0.005 \\
Value loss coefficient & 1.0 \\
Max gradient norm & 1.0 \\ \bottomrule
\end{tabular}
\end{table}

%% file: tables/deployable_obs.tex
\begin{table}[h]
\centering
\caption{Deployable Observations (Used by Tracker's Actor)}
\label{tab:deployable_obs}
\begin{tabular}{@{}lc@{}}
\toprule
\textbf{Observation Item} & \textbf{Dim.} \\ \midrule
History root angular velocity & 3 $\times$ 5 \\
History root joint position & 29 $\times$ 5 \\
History root joint velocity & 29 $\times$ 5 \\
History action & 29 $\times$ 5 \\
History root projected gravity & 3 $\times$ 5 \\
Object position & 3 \\
Object orientation & 6 \\

Command joint position & 29 \\
Command root linear velocity & 3 \\
Command root angular velocity & 3 \\
Command contact label & 1 \\

\bottomrule
\end{tabular}
\end{table}

%% file: tables/privileged_obs.tex
\begin{table}[h]
\centering
\caption{Privileged Observations (Used by Refiner's Actor/Critic and Tracker's Critic)}
\label{tab:privileged_obs}
\begin{tabular}{@{}lc@{}}
\toprule
\textbf{Observation Item} & \textbf{Dim.} \\ \midrule
Body position & 14 $\times$ 3 \\
Body orientation & 14 $\times$ 6 \\
Root linear velocity & 3 \\
Root angular velocity & 3 \\
Joint position & 29 \\
Joint velocity & 29 \\
Last action & 29 \\
Object position & 3 \\
Object orientation & 6 \\
Object linear velocity & 3 \\
Object angular velocity & 3 \\

Future reference joint position  & 29 $\times$ 5 \\
Future reference joint velocity  & 29 $\times$ 5 \\
Future reference root position & 3 $\times$ 5 \\
Future reference root orientation & 6 $\times$ 5 \\
Future reference object position & 3 $\times$ 5 \\
Future reference object orientation & 6 $\times$ 5 \\
Future reference object linear velocity & 3 $\times$ 5 \\
Future reference object angular velocity & 3 $\times$ 5 \\

\bottomrule
\end{tabular}
\end{table}

%% file: tables/reward.tex
\begin{table}[h]
\centering
\caption{Detailed Reward Terms and Hyperparameters. Here, $e$ denotes the error between the current state and the reference prior, $q$ represents joint positions, $\tau$ is the motor torque, and $a$ is the policy action. $\mathbb{I}(\cdot)$ is the indicator function.}
\label{tab:rewards}
\small
\begin{tabular}{@{}llcc@{}}
\toprule
\textbf{Category} & \textbf{Reward Term} & \textbf{Weight} & \textbf{Expression} \\ \midrule
\textbf{Tracking} & Joint Position & 0.125 & $\exp(-\|e\|^2 / 0.6^2)$ \\
 & Global Root Position & 0.25 & $\exp(-\|e\|^2 / 0.3^2)$ \\
 & Global Root Orientation & 0.25 & $\exp(-\|e\|^2 / 0.4^2)$ \\
 & Relative Body Position & 0.25 & $\exp(-\|e\|^2 / 0.3^2)$ \\
 & Relative Body Orientation & 0.25 & $\exp(-\|e\|^2 / 0.4^2)$ \\
 & Body Linear Velocity & 0.25 & $\exp(-\|e\|^2 / 1.0^2)$ \\
 & Body Angular Velocity & 0.25 & $\exp(-\|e\|^2 / 3.14^2)$ \\
 & Global Object Position & 0.5 & $\exp(-\|e\|^2 / 0.3^2)$ \\
 & Global Object Orientation & 0.5 & $\exp(-\|e\|^2 / 0.4^2)$ \\
 & Global Object Lin. Velocity & 0.5 & $\exp(-\|e\|^2 / 1.0^2)$ \\
 & Global Object Ang. Velocity & 0.5 & $\exp(-\|e\|^2 / 3.14^2)$ \\ \midrule
\textbf{Interaction} & Obj-to-Body Rel. Position & 0.25 & $\exp(-\|e\|^2 / 0.3^2)$ \\
 & Obj-to-Body Rel. Orientation & 0.25 & $\exp(-\|e\|^2 / 0.4^2)$ \\
 & Contact Consistency & 1.0 & $\mathbb{I}(F_{contact} > 1.0\text{N}) == \hat{l}_t$ \\ \midrule
\textbf{Regularization} & Feet Air Time & 5.0 & $\sum (t_{air} - 0.5)$ \\
 & Feet Slip Penalty & -0.1 & $\|v_{foot}^{xy}\| \cdot \mathbb{I}(F_{contact} > 1.0\text{N})$ \\
 & Undesired Contacts & -0.1 & $\mathbb{I}(F_{others} > 1.0\text{N})$ \\
 & Joint Acceleration & $-2.5 \times 10^{-7}$ & $\| \ddot{q} \|^2$ \\
 & Joint Torque & $-1.0 \times 10^{-5}$ & $\| \tau \|^2$ \\
 & Action Rate & -0.1 & $\| a_t - a_{t-1} \|^2$ \\
 & Joint Position Limits & -10.0 & $\mathbb{I}(q \notin [q_{min}, q_{max}])$ \\ \bottomrule
\end{tabular}
\end{table}

%% file: tables/domain_rand.tex
\begin{table}[h]
\centering
\caption{Domain Randomization Parameters}
\label{tab:domain_randomization}
\small
\begin{tabular}{@{}lc@{}}
\toprule
\textbf{Parameter} & \textbf{Range / Details} \\ \midrule
Robot static friction & [0.3, 1.6] \\
Robot dynamic friction & [0.3, 1.2] \\
Robot restitution & [0.0, 0.5] \\ 
Joint default position & $\pm 0.01$ rad \\ 
Base center of mass (x, y, z) & $\pm (0.025, 0.05, 0.05)$ m \\ \midrule
Object static friction & [0.2, 0.8] \\
Object dynamic friction & [0.2, 0.8] \\
Object restitution & [0.0, 0.5] \\ 
Object mass scale & [0.5, 2.0] \\ \midrule
Robot external push (lin, ang) & $\pm0.3$ m/s, $\pm0.5$rad/s \\
Object external push (lin, ang) & $\pm0.5$ m/s, $\pm1.0$rad/s \\
Push interval & [1.5, 3.0] s \\ \bottomrule
\end{tabular}
\end{table}

%% file: tables/early_termination.tex
\begin{table}[h]
\centering
\caption{Early Termination}
\label{tab:terminations}
\small
\begin{tabular}{@{}lc@{}}
\toprule
\textbf{Termination Term} & \textbf{Threshold} \\ \midrule
Root position error & 0.3 m \\
Root orientation error & 0.8 rad \\
Body position error & 0.3 m \\
Object position error & 0.3 m \\
Object orientation error & 0.8 rad \\ \bottomrule
\end{tabular}
\end{table}